# Semigraphoids are Two-Antecedental Approximations of Stochastic Conditional Independence Models

**Milan Studený***
Institute of Information Theory and Automation
Academy of Sciences of Czech Republic
Pod vodárenskou věží 4, 182 08 Prague, Czech Republic

## Abstract

The semigraphoid closure of every couple of CI-statements (CI=conditional independence) is a stochastic CI-model. As a consequence of this result it is shown that every probabilistically sound inference rule for CI-models, having at most two antecedents, is derivable from the semigraphoid inference rules. This justifies the use of semigraphoids as approximations of stochastic CI-models in probabilistic reasoning. The list of all 19 potential dominant elements of the mentioned semigraphoid closure is given as a byproduct.

## 1 INTRODUCTION

Many reasoning tasks arising in AI can be considerably simplified if a suitable concept of relevance or irrelevance of variables is utilized. The conditional irrelevance in probabilistic reasoning is modelled by means of the concept of stochastic *conditional independence* (CI) – details about the probabilistic approach to uncertainty handling in AI are in Pearl's book (1988). The fact that every CI-statement can be interpreted as certain qualitative relationship among symptoms or variables under consideration makes it possible to reduce the dimensionality of the problem and thus to find a more effective way of storing the knowledge base of a probabilistic expert system. This dimensionality reduction is important especially for the intensional approach developed by Perez and Jiroušek (1985). Nevertheless, the concept of CI has been introduced and studied for similar reasons also in other calculi of uncertainty (relational databases, Spohn's theory of ordinal conditional functions, possibility theory, Dempster–Shafer's theory) – see (Shenoy 1994), (Studený 1993).

The structures of CI were at first described by means of graphs in literature. Two trends can be recognized: by means of undirected graphs (the concept of the Markov net – see (Darroch *et al.* 1980), (Pearl 1988) and by means of directed acyclic graphs (the concepts of the Bayesian net (Pearl 1988), influence diagram (Shachter 1986), (Smith 1989) and recursive model (Kiiveri *et al.* 1984)). Nevertheless, both graphical approaches cannot describe all possible stochastic CI-structures. A natural way to remove this imperfection is to describe CI-structures by means of so-called *dependency models* i.e. lists of CI-statements. But such an approach would be unnecessarily wide as owing to well-known properties of stochastic CI (Dawid 1979), (Spohn 1980) many dependency models cannot be models of stochastic CI-structures. Therefore Pearl and Paz (1987) introduced the concept of the *semigraphoid* (resp. graphoid) as a dependency model closed under 4 (resp. 5) concrete inference rules[1] expressing the above mentioned properties of CI. Thus, every model of a stochastic CI-structure is a semigraphoid and Pearl (1988) conjectured the converse statement: that every semigraphoid is a CI-model (resp. every graphoid is a CI-model corresponding to a strictly positive measure).

Unfortunately, this conjecture was refuted firstly by finding a further sound and independent inference rule for stochastic CI-models (Studený 1989) and later even by showing that stochastic CI-models cannot be characterized by means of a finite number of inference rules (Studený 1992). This motivated an alternative approach to description of CI-structures by means of so-called imsets developed in (Studený 1994a). On the other hand, a finite characterization can be found for certain important substructures of CI-structure (Geiger and Pearl 1993).

However, all new independent inference rules for stochastic CI-models had more than two antecedents. Therefore Pearl and later also Paz during their visit in Prague formulated a hypothesis that all probabilistically sound inference rules for CI-models having at most two antecedents are in fact derivable from the semigraphoid inference rules i.e. that semigraphoids are "two-antecedental" approximations of stochastic CI-models. The new conjecture is confirmed in this paper by presenting a formally stronger result: the semigraphoid closure of every couple of CI-statements is a stochastic CI-model. As a corollary we obtain a characterization of this semigraphoid closure by means of potential dominant elements. As the proofs of these results are beyond of the scope of a conference contribution (they are in the paper (Studený 1994b) which has 26 pages) only definitions are given, results formulated and the ideas of proofs outlined in this paper.

*E-mail: studeny@utia.cas.cz

[1] Such an inference rule claims that if some CI-statements called *antecedents* are involved in a model of stochastic CI-structure then another CI-statement called the *consequent* has to be involved in that model of CI-structure, too.

## 2 BASIC DEFINITIONS AND FACTS

Throughout this paper the symbol $N$ denotes a finite nonempty set of *variables*. Every CI-statement will be described by means of a triplet $\langle A,B|C\rangle$[2] of pairwise disjoint subsets of $N$ where $A$ and $B$ are nonempty. The class of all such triplets will be denoted by $T(N)$. A *dependency model* is then simply a subset of $T(N)$.

The first step is to introduce stochastic CI-models:

<u>Definition 1</u> (CI-model, probabilistic implication)
A *probability measure* over $N$ is specified by a collection of nonempty finite sets $\{\mathbf{X}_i; i \in N\}$ and by a function $P: \Pi_{i\in N}\mathbf{X}_i \rightarrow [0,1]$ with $\sum\{P(x); x \in \Pi_{i\in N}\mathbf{X}_i\} = 1$.
Whenever $\emptyset \neq A \subset N$ and $P$ is a probability measure over $N$ its *marginal measure on* $A$ is a probability measure $P^A$ (over $A$) defined as follows ($P^N \equiv P$):
$P^A(a) = \sum\{P(a,b);\ b \in \Pi_{i\in N\setminus A}\mathbf{X}_i\} \qquad a \in \Pi_{i\in A}\mathbf{X}_i$.
Having $\langle A,B|C\rangle \in T(N)$ and a probability measure $P$ over $N$ we will say that CI-statement $A \perp B|C(P)$ is valid (i.e. $A$ is *conditionally independent of* $B$ *given* $C$ *with respect to* $P$) iff
$\forall\ a \in \Pi_{i\in A}\mathbf{X}_i \quad b \in \Pi_{i\in B}\mathbf{X}_i \quad c \in \Pi_{i\in C}\mathbf{X}_i$

$$P^{A\cup B\cup C}(a,b,c)\cdot P^C(c) = P^{A\cup C}(a,c)\cdot P^{B\cup C}(b,c)$$

(note the convention $P^{\emptyset}(-) = 1$).
Each probability measure $P$ over $N$ defines a dependency model over $N$ $\{\langle A,B|C\rangle \in T(N);\ A \perp B|C(P)\}$ called the *CI-model induced by P*. A dependency model is then called a *stochastic CI-model over N* iff it is induced by some probability measure over $N$.
Having $\{u_1,\ldots,u_k\} \subset T(N)$, $k \geq 1$ and $u_{k+1} \in T(N)$ we will say that $\{u_1,\ldots,u_k\}$ *probabilistically implies* $u_{k+1}$ and write $\{u_1,\ldots,u_k\} \models u_{k+1}$ iff <u>each</u> stochastic CI-model containing $\{u_1,\ldots,u_k\}$ also contains $u_{k+1}$.

Stochastic CI-models have the following important property – see (Geiger and Pearl 1990) or (Studený 1992) for the proof.

[2]Note that the order of components in the triplet used in this paper differs from (Pearl 1988), where the conditioned area is placed on the second position. We follow the original notation in probability theory: the conditioned area is on the third position after the separator $|$.

<u>Lemma 1</u>
The intersection of two stochastic CI-models over $N$ is also a stochastic CI-model over $N$.

As mentioned in the Introduction the purpose of the presented approach is to describe stochastic CI-models by means of so-called inference rules.

<u>Definition 2</u> (inference rule, semigraphoid)
By an *inference rule* with $r$ antecedents ($r \geq 1$) we will understand an $(r+1)$-nary relation on $T(N)$. We will say that a dependency model $M \subset T(N)$ is *closed under* an inference rule $\mathcal{R}$ iff for each *instance* of $\mathcal{R}$ (i.e. every collection $[u_1,\ldots,u_{r+1}]$ of elements of $T(N)$ belonging to $\mathcal{R}$) the following statement holds:
whenever the *antecedents* (i.e. $u_1,\ldots,u_r$) belong to $M$, then also the *consequent* (i.e. $u_{r+1}$) does so.
An inference rule $\mathcal{R}$ with $r$ antecedents is called (*probabilistically*) *sound* iff $\{u_1,\ldots,u_r\} \models u_{r+1}$ holds for every instance $[u_1,\ldots,u_{r+1}]$ of $\mathcal{R}$.
Usually, an inference rule is expressed by an informal schema, listening firstly antecedents and after an arrow the consequent. Thus, the following schemata :
$\langle A,B|C\rangle \rightarrow \langle B,A|C\rangle$ symmetry
$\langle A,B\cup C|D\rangle \rightarrow \langle A,C|D\rangle$ decomposition
$\langle A,B\cup C|D\rangle \rightarrow \langle A,B|C\cup D\rangle$ weak union
$[\langle A,B|C\cup D\rangle\ \&\ \langle A,C|D\rangle] \rightarrow \langle A,B\cup C|D\rangle$ contraction
describe four inference rules[3].
Every dependency model closed under these inference rules is called *semigraphoid*.
Moreover, we will say that $u_{k+1} \in T(N)$ is *derivable from* $\{u_1,\ldots,u_k\} \subset T(N)$ ($k \geq 1$) and write $\{u_1,\ldots,u_k\} \vdash_{sem} u_{k+1}$ iff there exists a derivation sequence $t_1,\ldots,t_n \subset T(N)$ where $t_n = u_{k+1}$ and for each $t_i$ ($i \leq n$) either $t_i \in \{u_1,\ldots,u_k\}$ or $t_i$ is a direct consequence of some preceding $t_j$s by virtue of some above mentioned semigraphoid inference rule.
Having a dependency model $M \subset T(N)$ its *semigraphoid closure* consists of all elements of $T(N)$ derivable from $M$ (it is evidently a semigraphoid).

As every semigraphoid inference rule is probabilistically sound it makes no problem to see:

<u>Lemma 2</u>
Whenever $u_1,\ldots,u_{k+1} \in T(N)$, then
$\{u_1,\ldots,u_k\} \vdash_{sem} u_{k+1}$ implies $\{u_1,\ldots,u_k\} \models u_{k+1}$.

## 3 MAIN RESULTS

Firstly, a special ordering on $T(N)$ is introduced.

<u>Definition 3</u> (dominating)
Supposing $\langle A,B|C\rangle, \langle X,Y|Z\rangle \in T(N)$ we will say that $\langle A,B|C\rangle$ *dominates* $\langle X,Y|Z\rangle$ and write
$\langle X,Y|Z\rangle \prec \langle A,B|C\rangle$ iff $\{X\cup Y\cup Z \subset A\cup B\cup C\}\ \&\ \{C \subset Z\}\ \&\ \{[X \subset A\ \&\ Y \subset B]$ or $[X \subset B\ \&\ Y \subset A]\}$.

[3]Some authors (for example (Geiger and Pearl 1993)) call inference rules of this type *Horn clauses*.

It is not difficult to see that the semigraphoid closure of one triplet $u \in T(N)$ (i.e. a singleton dependency model) consists of all triplets dominated by $u$. Also the proof that every such semigraphoid closure is a stochastic CI-model is simple. One can utilize the well-known result from (Geiger and Pearl 1990) that each Markov net defines a stochastic CI-model. Thus, given $u = \langle E, F|G \rangle$, first consider the graph with cliques $\{E \cup G, F \cup G\}$ and by the mentioned result find a ***strictly positive*** probability measure over $E \cup F \cup G$ whose CI-model is the set of triplets dominated by $u$. The second step is to find a strictly positive measure over $N$ having that prescribed marginal measure on $E \cup F \cup G$ and where no other CI-statement is valid. Such a 'conservative extension' can always be found for strictly positive measures – it is left as an excercise for the reader.

The second step is to find the potential dominant elements (i.e. maximal elements with respect to $\prec$) of the semigraphoid closure of a couple of elements of $T(N)$.

<u>Lemma 3</u>
Suppose that $\langle A, B|C \rangle$, $\langle I, J|K \rangle \in T(N)$ and $C \setminus (I \cup J \cup K) = \emptyset = K \setminus (A \cup B \cup C)$. Then each of the following 19 triplets belongs to the semigraphoid closure of $\{\langle A, B|C \rangle, \langle I, J|K \rangle\}$, of course provided that it is an element of $T(N)$ i.e. its first two components are nonempty sets.

1. $\langle A, B|C \rangle$
2. $\langle I, J|K \rangle$
3. $\langle A \cap I, (J \setminus C) \cup (B \cap (I \cup K))|C \cup (A \cap K) \rangle$
4. $\langle A \cap J, (I \setminus C) \cup (B \cap (J \cup K))|C \cup (A \cap K) \rangle$
5. $\langle B \cap I, (J \setminus C) \cup (A \cap (I \cup K))|C \cup (B \cap K) \rangle$
6. $\langle B \cap J, (I \setminus C) \cup (A \cap (J \cup K))|C \cup (B \cap K) \rangle$
7. $\langle A \cap I, (B \setminus K) \cup (J \cap (A \cup C))|K \cup (C \cap I) \rangle$
8. $\langle B \cap I, (A \setminus K) \cup (J \cap (B \cup C))|K \cup (C \cap I) \rangle$
9. $\langle A \cap J, (B \setminus K) \cup (I \cap (A \cup C))|K \cup (C \cap J) \rangle$
10. $\langle B \cap J, (A \setminus K) \cup (I \cap (B \cup C))|K \cup (C \cap J) \rangle$
11. $\langle A \cap I, B \cup (A \cap J)|C \cup (A \cap K) \rangle$
12. $\langle A \cap J, B \cup (A \cap I)|C \cup (A \cap K) \rangle$
13. $\langle B \cap I, A \cup (B \cap J)|C \cup (B \cap K) \rangle$
14. $\langle B \cap J, A \cup (B \cap I)|C \cup (B \cap K) \rangle$
15. $\langle A \cap J, I \cup (B \cap J)|K \cup (C \cap J) \rangle$
16. $\langle B \cap J, I \cup (A \cap J)|K \cup (C \cap J) \rangle$
17. $\langle A \cap I, J \cup (B \cap I)|K \cup (C \cap I) \rangle$
18. $\langle B \cap I, J \cup (A \cap I)|K \cup (C \cap I) \rangle$
19. $\langle (A \cap I) \cup (B \cap J), (A \cap J) \cup (B \cap I)|C \cup K \rangle$

The potential dominant elements mentioned in Lemma 3 are shown in the following figures. In the figures, the set conditioned on is in black and the independent areas are shown by vertical and horizontal hatching, respectively.

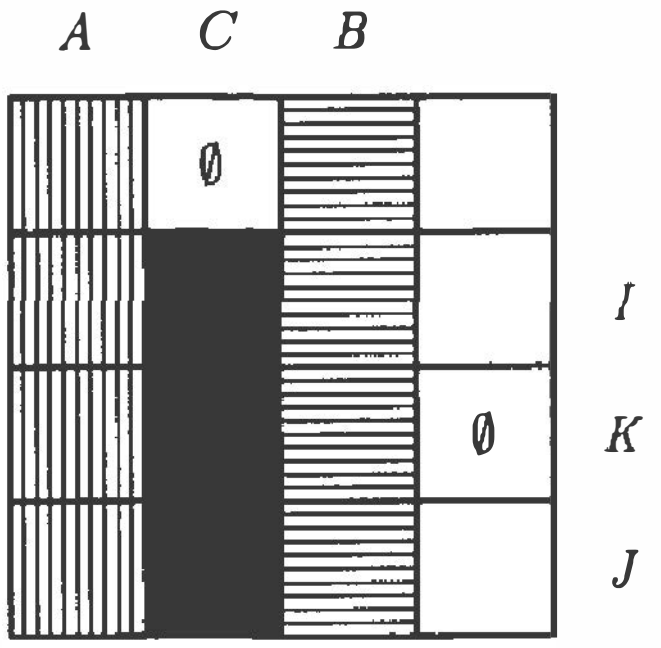

Figure 1: Triplet n. 1

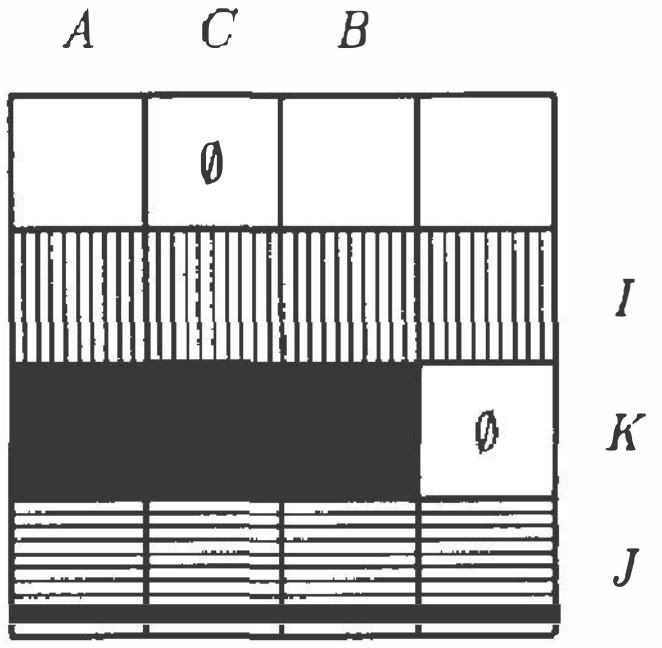

Figure 2: Triplet n. 2

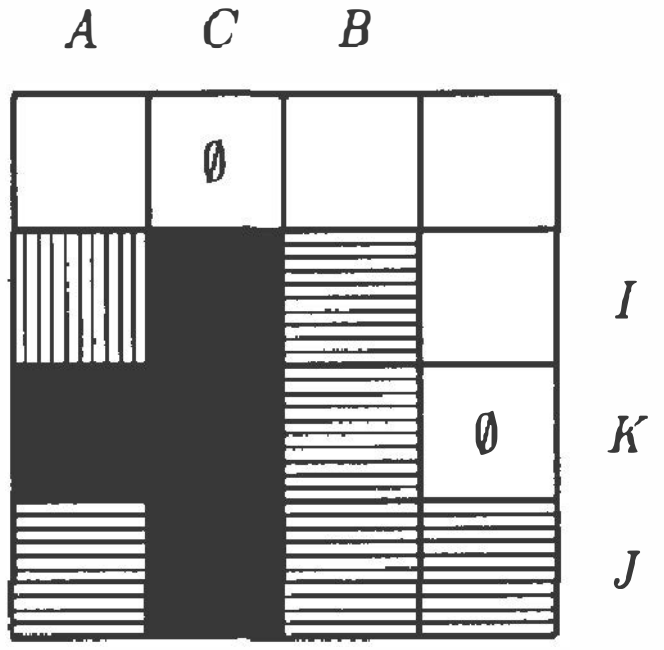

Figure 3: Triplet n. 3

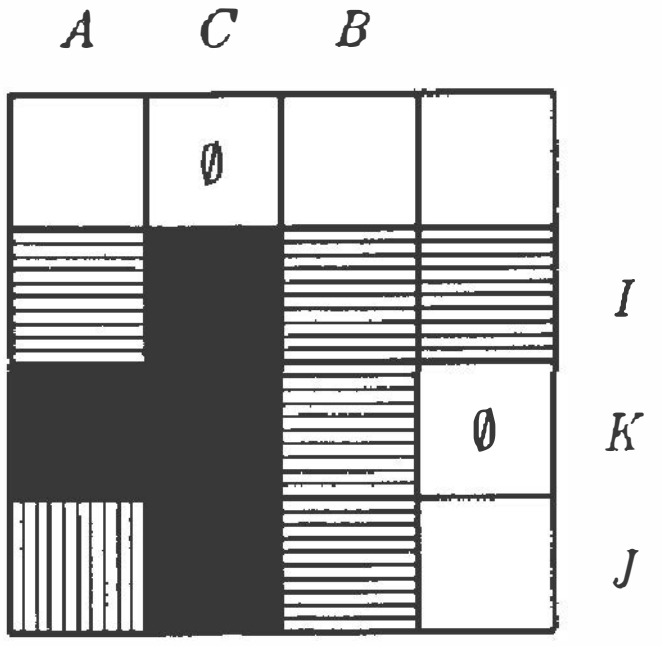

Figure 4: Triplet n. 4

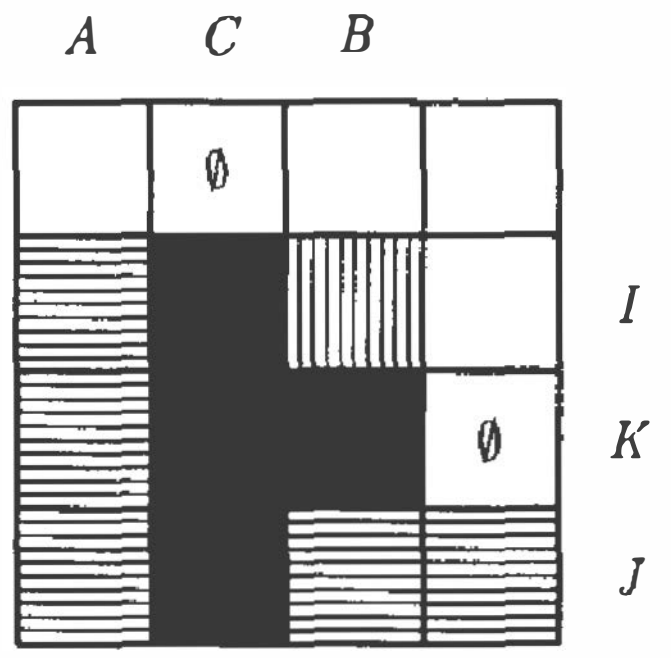

Figure 5: Triplet n. 5

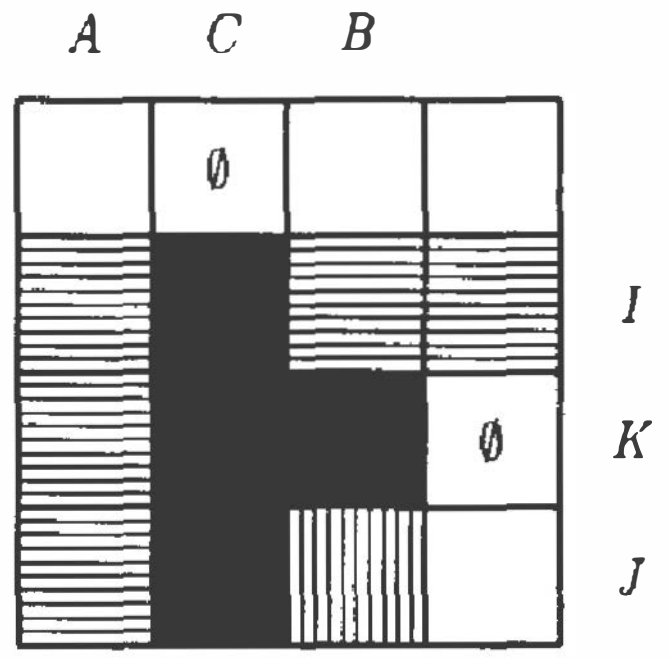

Figure 6: Triplet n. 6

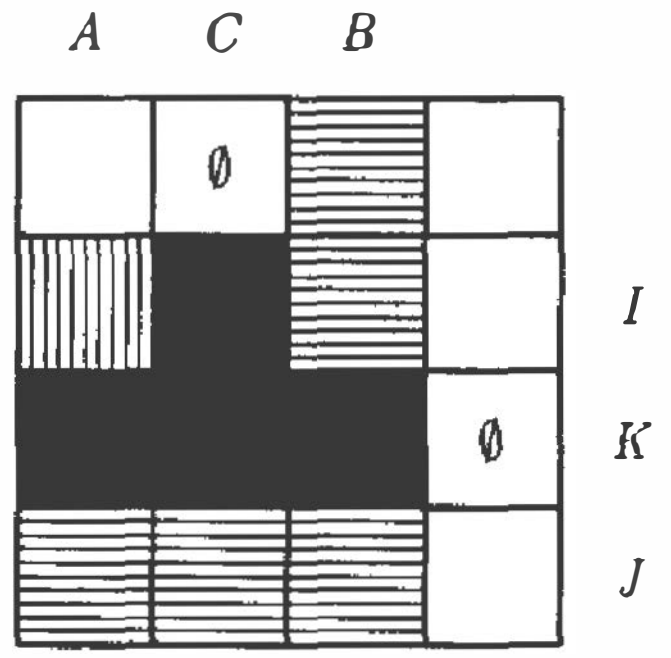

Figure 7: Triplet n. 7

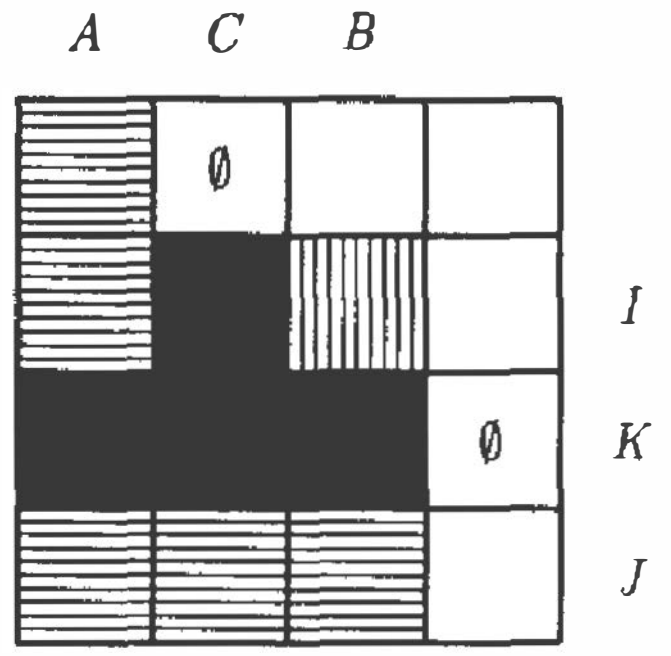

Figure 8: Triplet n. 8

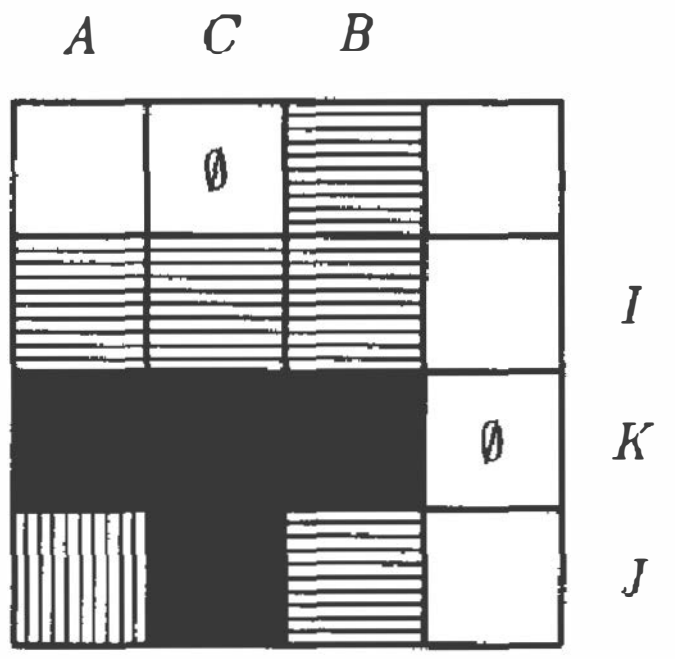

Figure 9: Triplet n. 9

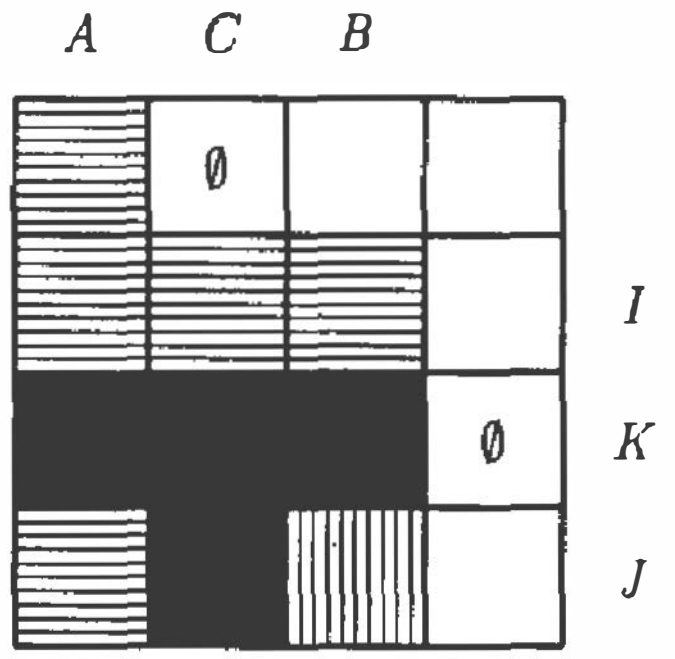

Figure 10: Triplet n. 10

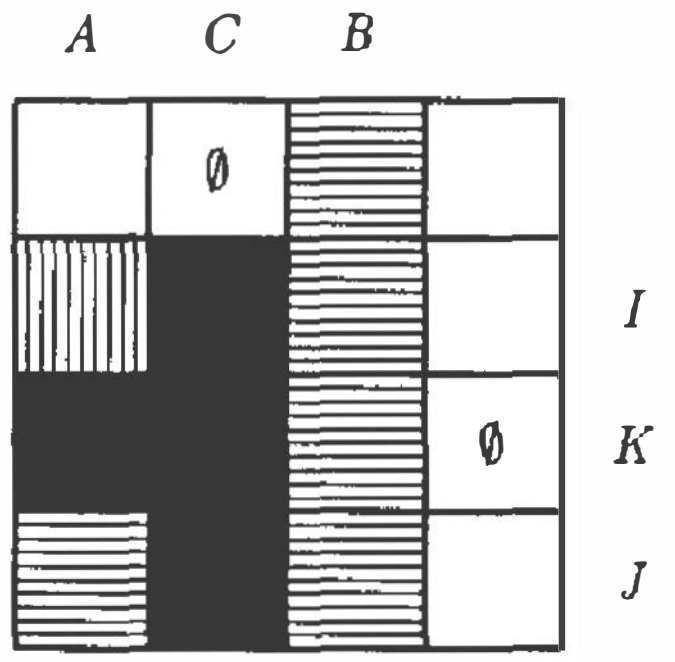

Figure 11: Triplet n. 11

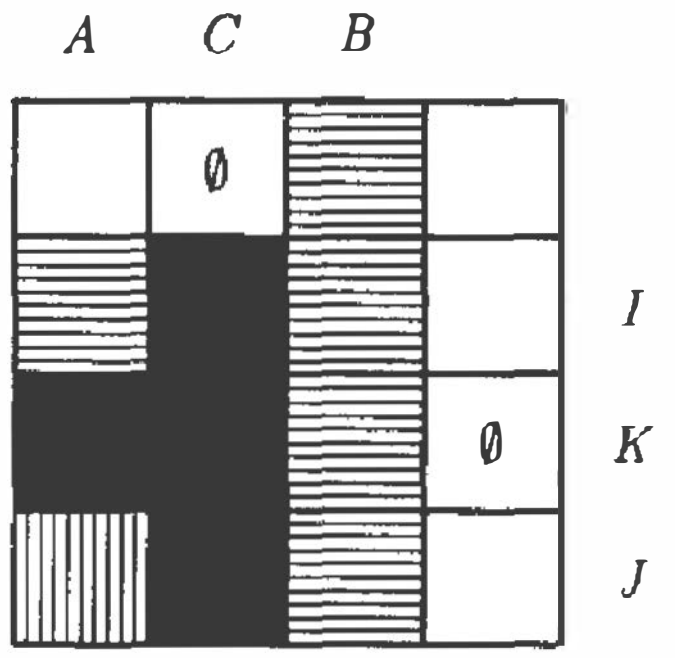

Figure 12: Triplet n. 12

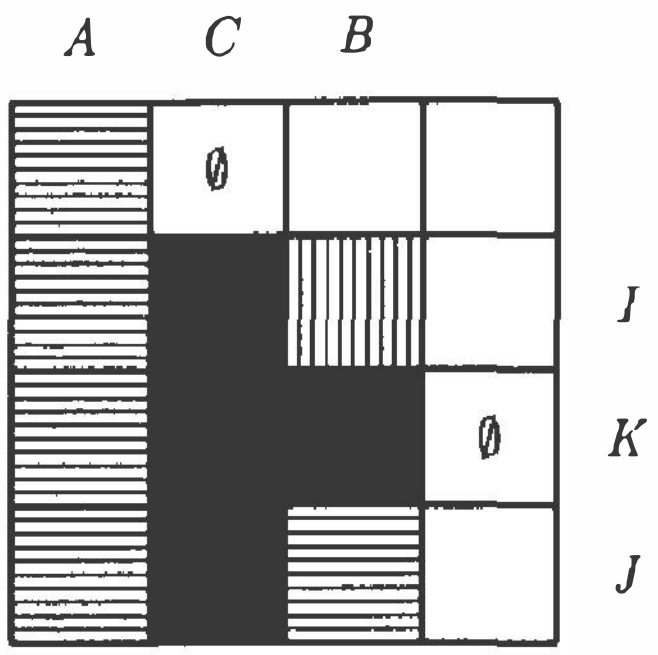

Figure 13: Triplet n. 13

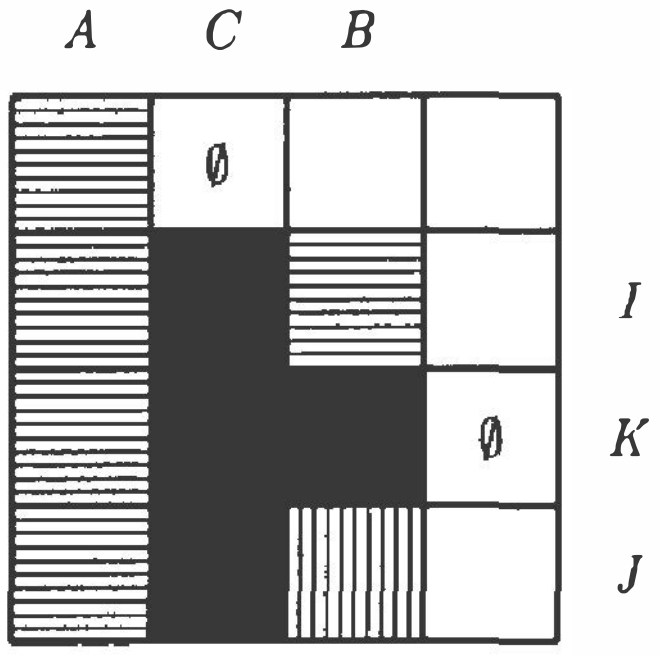

Figure 14: Triplet n. 14

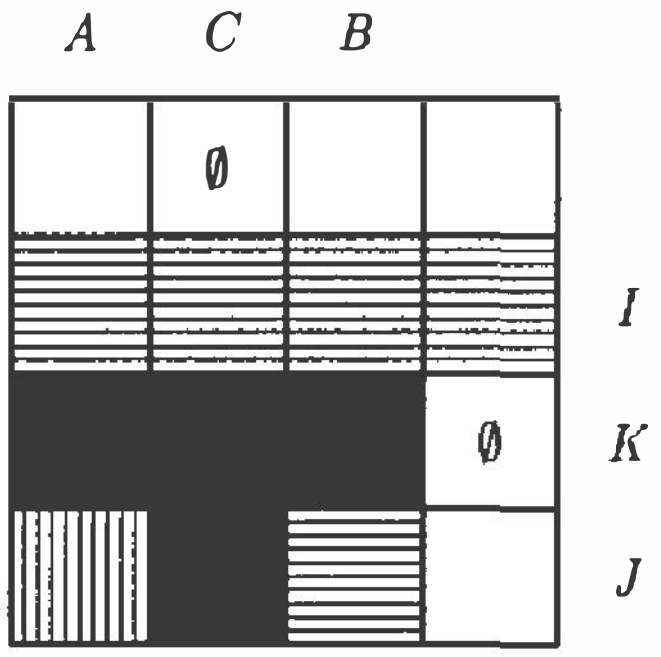

Figure 15: Triplet n. 15

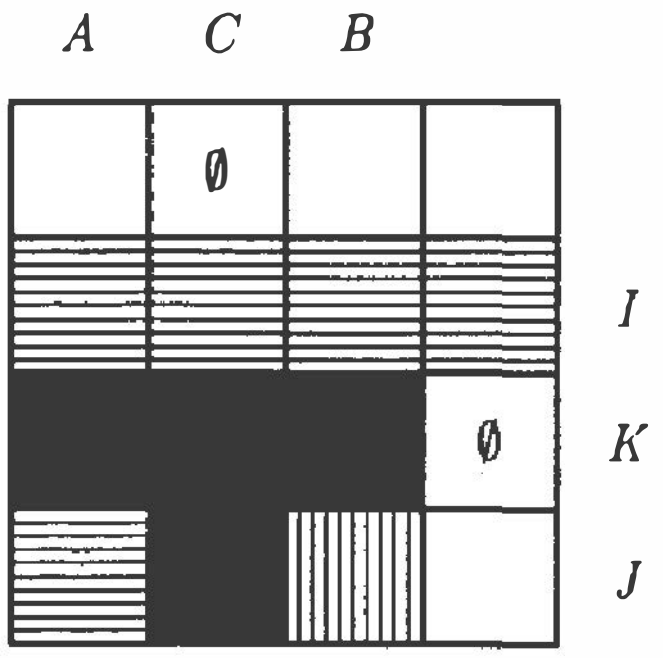

Figure 16: Triplet n. 16

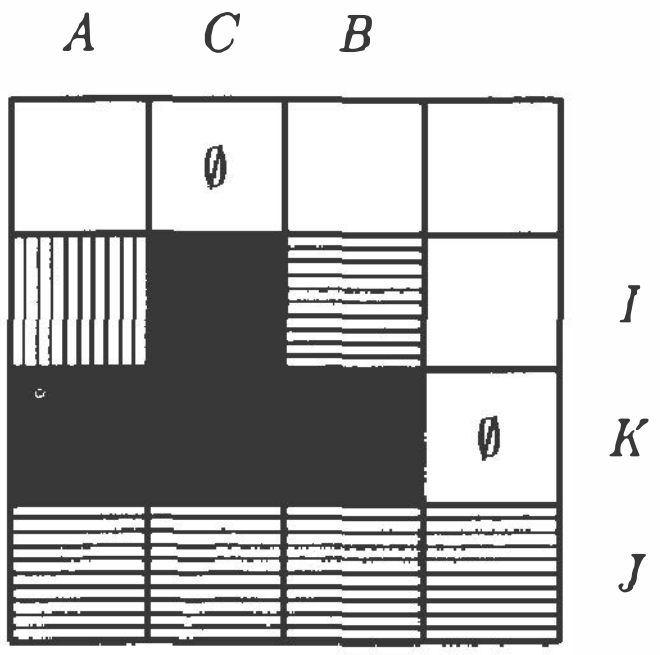

Figure 17: Triplet n. 17

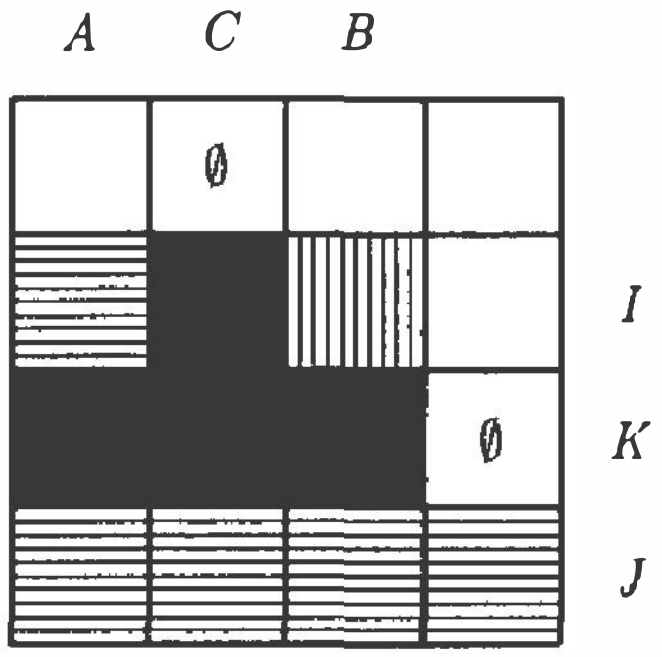

Figure 18: Triplet n. 18

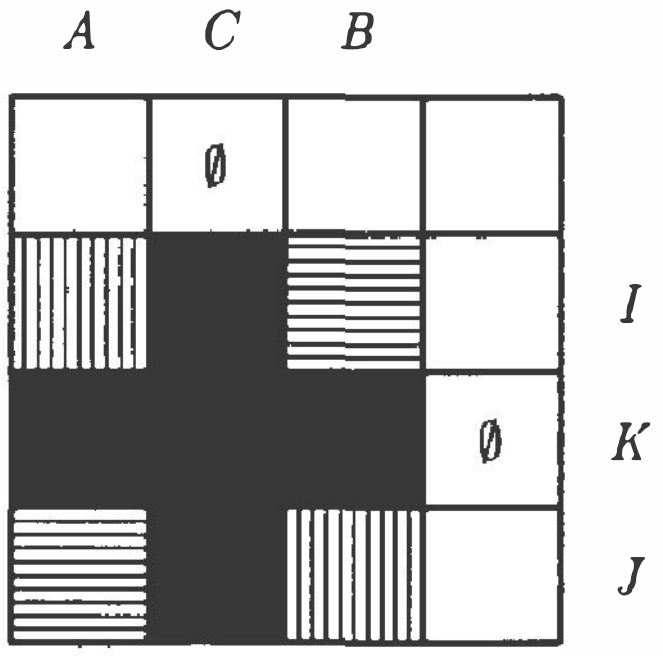

Figure 19: Triplet n. 19

*Proof of Lemma 3:* Owing to the permutations $I \leftrightarrow J$, $A \leftrightarrow B$, $[A, B, C] \leftrightarrow [I, J, K]$ it suffices to derive the triplets n. 3, 11, 19 only. The triplet n. 3 is derived by virtue of contraction from $\langle A \cap I, B \cap (I \cup K) | C \cup (A \cap K) \rangle \prec \langle A, B | C \rangle$ and $\langle A \cap I, J \setminus C | (B \cap (I \cup K)) \cup C \cup (A \cap K) \rangle \prec \langle I, J | K \rangle$. The triplet n. 11 is derived by virtue of contraction from $\langle A \cap I, B | (A \cap J) \cup C \cup (A \cap K) \rangle \prec \langle A, B | C \rangle$ and $\langle A \cap I, (A \cap J) | C \cup (A \cap K) \rangle$ which is dominated by the triplet n. 3. Finally, the triplet n. 19 is derived from $\langle (A \cap I) \cup (B \cap J), A \cap J | (B \cap I) \cup C \cup K \rangle$ and $\langle (A \cap I) \cup (B \cap J), B \cap I | C \cup K \rangle$ which are dominated by the triplet n. 4 respectively by the triplet n. 5. □

Then in (Studený 1994b) ten special constructions of probability measures are given. Concretely, we need 1 twodimensional construction (i.e. involving two random variables), 2 threedimensional constructions, 5 fourdimensional constructions and 2 fivedimensional constructions. These constructions allow us to show that supposing $\{\langle A,B|C\rangle,\ \langle I,J|K\rangle\} \models \langle X,Y|Z\rangle$ many specific situations (their number is about 60) are impossible, for example: $Z\setminus(A\cup B\cup C\cup I\cup J\cup K) \neq \emptyset$ or $[X\cap J \neq \emptyset \ \&\ Y\cap J \neq \emptyset \ \&\ (C\cap J)\setminus(X\cup Y\cup Z) \neq \emptyset]$ etc. These arguments are then used to prove (the most technical part of (Studený 1994b)):

Lemma 4
Suppose that $\langle A,B|C\rangle, \langle I,J|K\rangle, \langle X,Y|Z\rangle \in T(N)$ and $\{\langle A,B|C\rangle, \langle I,J|K\rangle\} \models \langle X,Y|Z\rangle$.
a) If $C\setminus(I\cup J\cup K) \neq \emptyset$ or $K\setminus(A\cup B\cup C) \neq \emptyset$, then the triplet $\langle X,Y|Z\rangle$ is dominated either by $\langle A,B|C\rangle$ or by $\langle I,J|K\rangle$.
b) If $[C\setminus(I\cup J\cup K) = \emptyset \ \&\ K\setminus(A\cup B\cup C) = \emptyset]$, then the triplet $\langle X,Y|Z\rangle$ has to be dominated by one of 19 triplets listened in Lemma 3.

Thus, by Lemmas 2, 3, 4 we easily get :

$$[\{\langle A,B|C\rangle,\langle I,J|K\rangle\} \models \langle X,Y|Z\rangle] \Leftrightarrow \Leftrightarrow [\{\langle A,B|C\rangle,\langle I,J|K\rangle \vdash_{sem} \langle X,Y|Z\rangle]$$

for $\langle A,B|C\rangle$, $\langle I,J|K\rangle$, $\langle X,Y|Z\rangle \in T(N)$ and

Corollary 1
Supposing $\langle A,B|C\rangle$, $\langle I,J|K\rangle \in T(N)$ let $M \subset T(N)$ be the semigraphoid closure of $\{\langle A,B|C\rangle,\ \langle I,J|K\rangle\}$. Then $M$ can be characterized as follows.
a) If $C\setminus(I\cup J\cup K) \neq \emptyset$ or $K\setminus(A\cup B\cup C) \neq \emptyset$, then $u \in M$ iff $u$ is dominated either by $\langle A,B|C\rangle$ or by $\langle I,J|K\rangle$.
b) If $[C\setminus(I\cup J\cup K) = \emptyset \ \&\ K\setminus(A\cup B\cup C) = \emptyset]$, then $u \in M$ iff $u$ is dominated by one of 19 potential dominant elements listened in Lemma 3.
Especially, $M$ has at most 19 dominant elements of $T(N)$.

Moreover, using Lemma 1 the main result can be shown:

THEOREM
The semigraphoid closure of a couple of elements of $T(N)$ (i.e. of a dependency model of cardinality 2) is a stochastic CI-model.

*Proof:* Supposing $u,v \in T(N)$ let $M \subset T(N)$ be the semigraphoid closure of $\{u,v\}$. For each $t \in T(N)\setminus M$ we have $\{u,v\} \not\vdash_{sem} t$ what is equivalent to $\{u,v\} \not\models t$. Thus, one can find a CI-model $I_t$ with $t \notin I_t \supset \{u,v\}$. By Lemma 1 $M = \bigcap_{t\in T(N)\setminus M} I_t$ is a CI-model. □

Besides, a further consequence can be derived by a simple consideration:

Corollary 2
Every probabilistically sound inference rule with at most two antecedents is derivable from the semigraphoid inference rules.

*Proof:* Suppose that $\mathcal{R}$ is such an inference rule (for example with two antecedents), let $[u_1,u_2,u_3] \in \mathcal{R}$ be its instance. By probabilistic soundness of $\mathcal{R}$ we have $\{u_1,u_2\} \models \{u_3\}$. Hence $\{u_1,u_2\} \vdash_{sem} u_3$ i.e. $u_3$ is derivable from $\{u_1,u_2\}$ by the semigraphoid inference rules. As every instance of $\mathcal{R}$ is already covered by the semigraphoid derivability the whole rule $\mathcal{R}$ is unnecessary. □

# 4 CONCLUSION

The result presented in this paper has theoretical significance for one of fundamental calculi for uncertainty handling in AI – for probabilistic reasoning. It confirms Pearl's and Paz's revised conjecture that the only sound inference rules for stochastic CI-models having at most two antecedents are those derivable from the semigraphoid inference rules. Thus, it shows that the concept of semigraphoid is at least partially justified and Pearl's original conjecture was well-founded. One can interpret this result by saying that semigraphoids are "two-antecedental" approximations of stochastic CI-models. Note for information that the only independent[4] sound inference rule with 3 antecedents known so far for stochastic CI-models is the following one – see (Studený 1992):

$$[\langle A,B|C\rangle \ \&\ \langle A,C|D\rangle \ \&\ \langle A,D|B\rangle] \ \rightarrow\ \langle A,C|B\rangle.$$

Nevertheless, our main result may have also more concrete significance. One can sometimes face the task to show that a given dependency model is a stochastic CI-model. The presented theorem gives a relatively simple sufficient condition: every semigraphoid closure of a couple of CI-statements is a stochastic CI-model. The characterization of this semigraphoid closure given in Corollary 1 is then beneficial.

**Acknowledgements**

I am grateful to both reviewers for helpful comments. This work was supported by the internal grants of the Academy of Sciences of Czech Republic n. 27564 "Knowledge derivation for probabilistic expert systems" and n. 275105 "Conditional independence properties in uncertainty processing".

[4] i.e. not derivable from the semigraphoid inference rules